%% file: main.tex
\documentclass[10pt,twocolumn,letterpaper]{article}

\usepackage[pagenumbers]{cvpr} % To force page numbers, e.g. for an 

\input{preamble}

\definecolor{cvprblue}{rgb}{0.21,0.49,0.74}
\usepackage[pagebackref,breaklinks,colorlinks,allcolors=cvprblue]{hyperref}

\def\paperID{*****} % *** Enter the Paper ID here
\def\confName{CVPR}
\def\confYear{2026}

\title{DDiT: Dynamic Patch Scheduling for Efficient Diffusion Transformers}

\author{
Dahye Kim$^{1,2}$\thanks{Work done during internship at Amazon}\footnotemark[1]\quad\quad
Deepti Ghadiyaram$^{1}$\quad\quad
Raghudeep Gadde$^{2}$ \\
$^1$Boston University \quad
$^2$Amazon \\
{\tt\small {\{dahye, dghadiya\}@bu.edu} \quad\quad  raghudeep.g@gmail.com}}

\begin{document}
\makeatletter
\begingroup
\renewcommand\thefootnote{\fnsymbol{footnote}}
\setcounter{footnote}{0}

\twocolumn[
\maketitle
\vspace{-2em}
\input{figs_latex/fig1}
\vspace{1em}
]
\footnotetext[1]{Work done as an intern at Amazon.}

\endgroup
\makeatother

\input{sec/0_abstract}    
\vspace{-0.16in}
\input{sec/1_intro}

\input{sec/2_related_work}

\input{sec/3_method}

\input{sec/4_experiments}

\input{sec/5_conclusion}

\clearpage
{
    \small
    \bibliographystyle{ieeenat_fullname}
    \bibliography{main}
}

\end{document}

%% file: preamble.tex
\newcommand{\modelname}{\textit{DDiT}}

\usepackage[normalem]{ulem}
\usepackage{xcolor}         % colors
\usepackage{tcolorbox}

\usepackage{algorithm}
\usepackage{algorithmic}
\usepackage{multirow} 
\usepackage{float} 
\usepackage[table]{xcolor} 
\usepackage{tabularx}
\usepackage{booktabs}
\usepackage{amsmath} % For math environments and symbols
\usepackage{amssymb} % For additional math symbols
\usepackage{booktabs} % For professional-looking tables (toprule, midrule, bottomrule)
\usepackage{tcolorbox}
\usepackage{listings}
\usepackage{caption}

%% file: figs_latex/fig1.tex
\begin{center}
\includegraphics[width=0.99\linewidth]{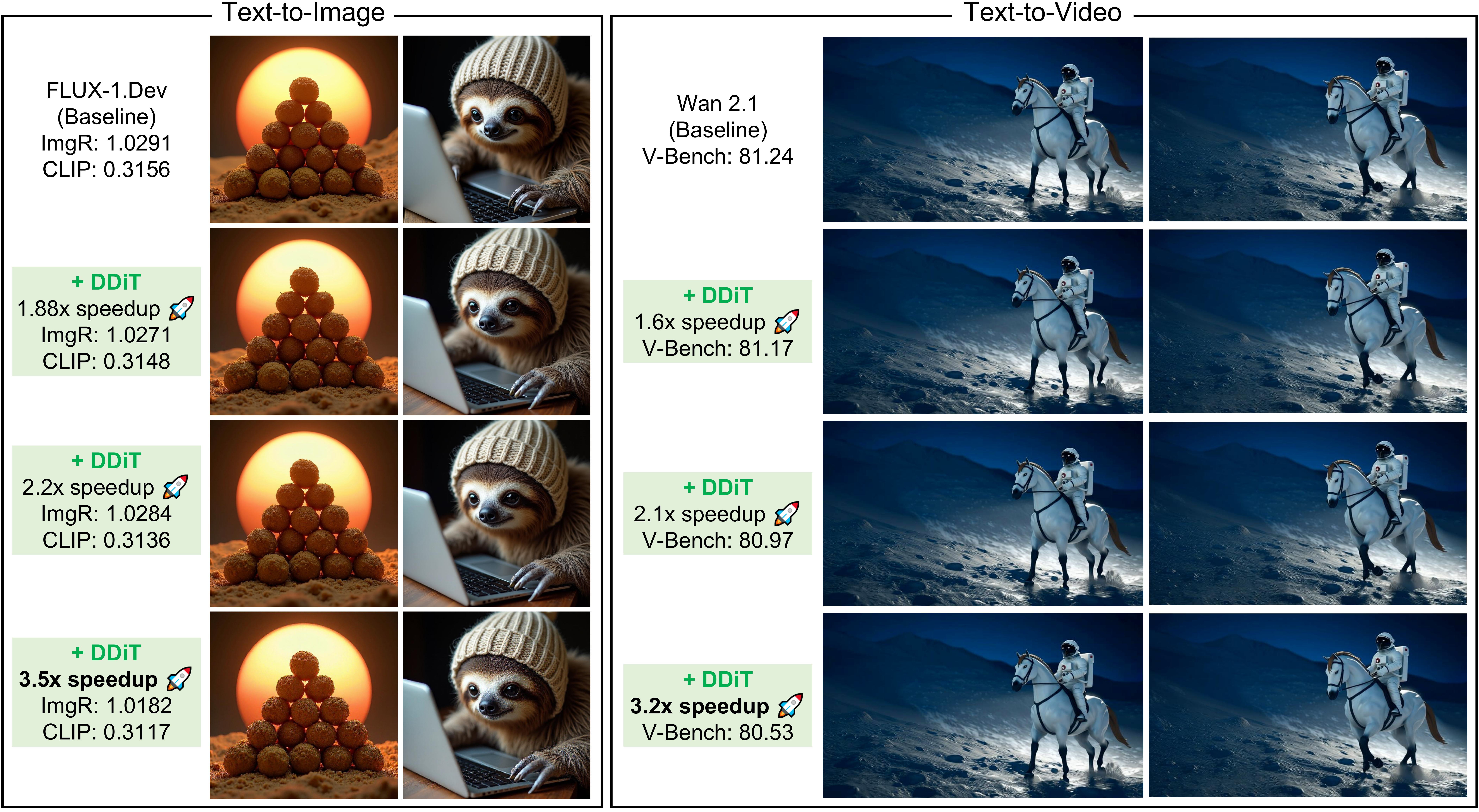}\vspace{-8pt}
\captionsetup{font=footnotesize}
  \captionof{figure}{\textbf{DDiT dynamically selects the optimal patch size} at each denoising step at inference yielding significant computational gains at no loss of perceptual quality. Results are shown for FLUX-1.Dev~\cite{flux2023} for text-to-image and {Wan-2.1}~\cite{wan2025} for text-to-video generation. 
The top panel denotes the baseline (original model), while the remaining panels illustrate outputs from DDiT at different acceleration rates. 
{ImageReward}~\cite{xu2023imagereward}, {CLIP}~\cite{radford2021learning}, and {VBench}~\cite{huang2024vbench} scores are reported (higher is better). 
}\vspace{-6pt}
  \label{fig:overview}
\end{center}

%% file: sec/0_abstract.tex
\begin{abstract}
Diffusion Transformers (DiTs) have achieved state-of-the-art performance in image and video generation, but their success comes at the cost of heavy computation. This inefficiency is largely due to the fixed tokenization process, which uses constant-sized patches throughout the entire denoising phase, regardless of the content's complexity.
We propose \emph{dynamic tokenization}, an efficient \textbf{test-time strategy} that varies patch sizes based on content complexity and the denoising timestep. Our key insight is that early timesteps only require coarser patches to model global structure, while later iterations demand finer (smaller-sized) patches to refine local details. During inference, our method dynamically reallocates patch sizes across denoising steps for image and video generation and substantially reduces cost while preserving perceptual generation quality. Extensive experiments demonstrate the effectiveness of our approach: it achieves up to $3.52\times$ and $3.2\times$ speedup on FLUX-1.Dev and Wan $2.1$, respectively, without compromising the generation quality and prompt adherence.
\end{abstract}

%% file: sec/1_intro.tex
\section{Introduction}
\label{sec:intro}

Diffusion transformers (DiTs)~\cite{peebles2023scalable, esser2024scaling, flux2023,seedream2025seedream, cai2025hidream, kong2024hunyuanvideo, wu2025qwen} have emerged as a dominant framework for content generation, producing high-quality and photorealistic results in both image and video synthesis. 
These advances have facilitated a wide range of applications, including image and video editing~\cite{brooks2023instructpix2pix, kawar2023imagic}, subject-driven generation~\cite{ruiz2023dreambooth, wang2024omnicontrolnet}, and digital art creation~\cite{mazzone2019art}. However, this impressive performance comes with substantial computational cost -- generating a \textit{single} $5$ second $720p$ video using Wan-2.1~\cite{wan2025} on an RTX 4090 takes $30$ minutes! -- significantly limiting the usage of these models in practice.
Such high computational demands of generative models have catapulted the development of more efficient generation methods. Existing research has broadly focused on acceleration techniques such as feature caching
~\cite{ma2024learning, zhang2025blockdance, liu2025timestep, liu2025reusing}, feature pruning
~\cite{bolya2023token, fang2023structural, wang2024attention, zhang2024laptop}, vector quantization~\cite{shang2023post, so2023temporal, tian2024qvd, deng2025vq4dit}, and model distillation~\cite{salimans2022progressive, li2023snapfusion, kim2024bk, zhang2024accelerating}.

Although these approaches show promise, they suffer from two key limitations. First, many methods~\cite{hu2024token, fang2025tinyfusion, wang2023patch,ma2024learning} typically employ a \textbf{hard, static reduction strategy}, such as removing a fixed amount of weights, operations, or tokens.
Such static approach can lead to significant quality degradation, as computations critical to a specific output might be permanently discarded~\cite{wang2023patch, deng2025vq4dit}. 
% \deepti{cite an example from any of hte above papers}
Second, most existing methods~\cite{ma2024deepcache, wang2023patch, ma2024learning}
% \deepti{cite which ones} 
apply a rigid, one-size-fits-all strategy, that is agnostic to the input. This is problematic, as different prompts require varying levels of computational detail~\cite{wang2023not, mahajan2024prompting}. A simple prompt like ``a blue sky'' should not require the same amount of computational resources compared to a prompt ``a scene crowded with many zebras.'' The rigidity in all existing solutions prevents us from dynamically allocating resources where they are needed most.

In this work, we address the rigid, one-size-fits-all computation of existing methods. Our approach is based on a key observation: the visual content generated by a diffusion model evolves at varied levels of detail. Some denoising timesteps establish coarse scene structure, while others refine fine-grained visual details.
Recent studies~\cite{li2023your, stracke2025cleandift, tang2023emergent, kim2025revelio} show that features generated at different timesteps of the denoising process encode different information, thus selecting the right timestep of diffusion features is important for successful downstream tasks such as classification~\cite{li2023your}, visual reasoning~\cite{kim2025revelio}, visual correspondence~\cite{tang2023emergent}, and semantic segmentation~\cite{stracke2025cleandift}.
Furthermore, ~\cite{patashnik2023localizing, wang2023not} note that this information can also be used for image editing, catering to different levels of detail in an image~\cite{mahajan2024prompting, patashnik2023localizing, wang2023not}, and for generating more prompt-aligned images by injecting different levels of prompt information at different timesteps~\cite{saichandran2025progressive}.

This leads us to a critical question: \textbf{should every denoising step process the latent at the same granularity?} Or, could some steps operate on a coarser latent, thereby yielding computational benefits, while others use a finer latent to preserve detail?
Thus, unlike prior works~\cite{bolya2023token, fang2023structural, wang2024attention, zhang2024laptop} which approach efficiency by discarding weights or operations, we \textit{dynamically allocate} it. Specifically, at every timestep, we adjust the patch size of the latent and adaptively use larger patches (coarser granularity) when less detail is required and smaller patches (finer granularity) when high fidelity is needed.

This, however, raises a new question: \textbf{how do we determine the optimal patch size at any given timestep and for any given prompt?} For this, we measure the \textit{rate of change of the latent manifold} over time. We hypothesize that this rate correlates with the level of detail being generated. If the underlying latent evolves slowly within a short timestep window, we posit that coarse-grained details are being generated. Consequently, we divide the latent into coarser patches and process them, saving computational resources. Conversely, if the underlying latent evolves rapidly, we infer that fine-grained details are being generated and fall back to using finer-grained latent patches.

Thus, this dynamic strategy tailors the computation load to each timestep and each prompt, allocating more resources when needed and conserving them where possible. Ultimately, our approach gives us explicit control over the computational budget while generating the highest possible quality content given the computational budget (Fig.~\ref{fig:overview}). In summary, our contributions are:

\begin{itemize}
\item We introduce a simple, intuitive, and low-cost strategy to \textbf{dynamically vary latent's granularity} in diffusion models, that requires minimal architectural changes (Fig.~\ref{fig:fig2}).
\item We propose a test-time \textbf{Dynamic Patch Scheduler} that automatically determines the optimal patch size at each timestep, adapting the computational load based on generation complexity and the input prompt.
\item We demonstrate through extensive experiments that our approach generalizes across both image and video diffusion transformers and achieves significant speedups -- up to $3.52\times$ on FLUX-1.Dev~\cite{flux2023} and $3.2\times$ speedups on Wan~2.1~\cite{wan2025}, while maintaining high perceptual quality, photo-realism, and prompt alignment.
\item We provide a detailed analysis of the rate of latent manifold evolution to generative complexity, offering a new perspective on the internal dynamics of diffusion models.
\end{itemize}

%% file: sec/2_related_work.tex
\section{Related work}
\label{sec:related_work}
\input{figs_latex/fig2}

\noindent\textbf{Efficient diffusion transformers.}
Diffusion transformers incur substantial computational costs due to their iterative denoising and attention operations. 
To address this challenge, a growing body of work has focused on improving the efficiency of these models through various algorithmic and architectural strategies.
Fast sampling methods~\cite{song2020denoising, lu2022dpm, lu2025dpm, zheng2023dpm, liu2022flow, meng2023distillation, salimans2022progressive, yang2023weighted, park2024textit, zhang2023adadiff,fernandez2024duodiff, nguyen2025swiftedit, gwilliam2025accelerate} reduce the number of sampling steps while preserving output quality.
Caching-based methods~\cite{ma2024learning, fang2025attend, liu2025timestep, liu2025reusing, li2023faster, ma2024deepcache, chen2024delta, selvaraju2024fora, wimbauer2024cache, lou2024token, huang2024harmonica, lv2024fastercache, kahatapitiya2025adaptive, gao2024ca2, yuan2024car, zou2024accelerating, saghatchian2025cached, zhang2024token, liu2025timestep, ye2024training, sun2025unicp, liu2025region, zhang2025blockdance} improve efficiency by reusing previously computed intermediate representations to avoid redundant computation.
Pruning-based methods~\cite{bolya2023token, fang2023structural, wang2024attention, zhang2024laptop, wang2024sparsedm, kim2024token,lou2024token, zhao2024dynamic, smith2024todo, ju2023turbo, su2024f3, zhu2024dip, tian2024u, xu2024headrouter, avrahami2025stable, fang2025tinyfusion, kim2024diffusion, chang2024flexdit, sun2024asymrnr, lee2024koala,lu2025toma, hu2024token,you2025layer, saghatchian2025cached, zhang2024token, cheng2025cat, singh2024negative, sun2025unicp} accelerate inference by removing redundant or less informative model weights, thereby reducing the number of operations.
Quantization-based methods~\cite{shang2023post, so2023temporal, tian2024qvd, deng2025vq4dit,dong2025ditas, chen2025q, li2024svdquant, fan2025sq} improve efficiency by converting model weights and activations from high-precision to low-precision representations, such as 8-bit integers~\cite{dettmers2023qlora}.
Knowledge distillation methods~\cite{salimans2022progressive, li2023snapfusion, kim2024bk, zhang2024accelerating, feng2024relational, zhu2024accelerating, chen2025snapgen, park2025inference} achieve efficiency by compressing complex models into smaller version using distillation objectives~\cite{hinton2015distilling}.
Although these approaches have shown promising results in reducing computation, they typically rely on hard, predefined reduction rules that lack adaptivity to content complexity.
Such hard constraints often discard essential details or oversimplify fine structures, ultimately degrading generation quality.
In contrast, we dynamically allocate computation across timesteps for efficient yet high-quality generation. \\
\noindent\textbf{Dynamic patch sizing for efficient transformers.}
Several prior works have explored using multiple patch sizes in transformer-based architectures.  
Methods such as~\cite{chen2021crossvit, beyer2023flexivit, wang2024multi, wang2021not, zhou2023make, hu2024lf} train models capable of operating with different patch sizes across images in ViTs.  
To further enhance efficiency, subsequent approaches~\cite{an2024sharpose, ronen2023vision, chen2023cf, bai2024coarse, choudhury2025accelerating} enable adaptive patch sizes within a single image, allowing the model to allocate computation based on local content complexity.  
Similarly, several works have investigated using different patch sizes or resolutions in DiTs~\cite{ho2022cascaded, saharia2022photorealistic, ho2022imagen, saharia2022image, pernias2023wurstchen, gu2023matryoshka, chang2024flexdit, atzmon2024edify, jin2024pyramidal, chen2025pixelflow}.  
However, all of these methods either 1) require training from scratch with sophisticated architectural designs, 2) are not generalizable to existing off-the-shelf pretrained DiTs, or 3) use a rigid and manually defined schedule for patch size during inference.  
We propose \modelname, a generic framework that dynamically adjusts patch sizes during test time for efficient generation.

%% file: figs_latex/fig2.tex
\begin{figure*}[t]
  \centering
   \includegraphics[width=0.88\linewidth]{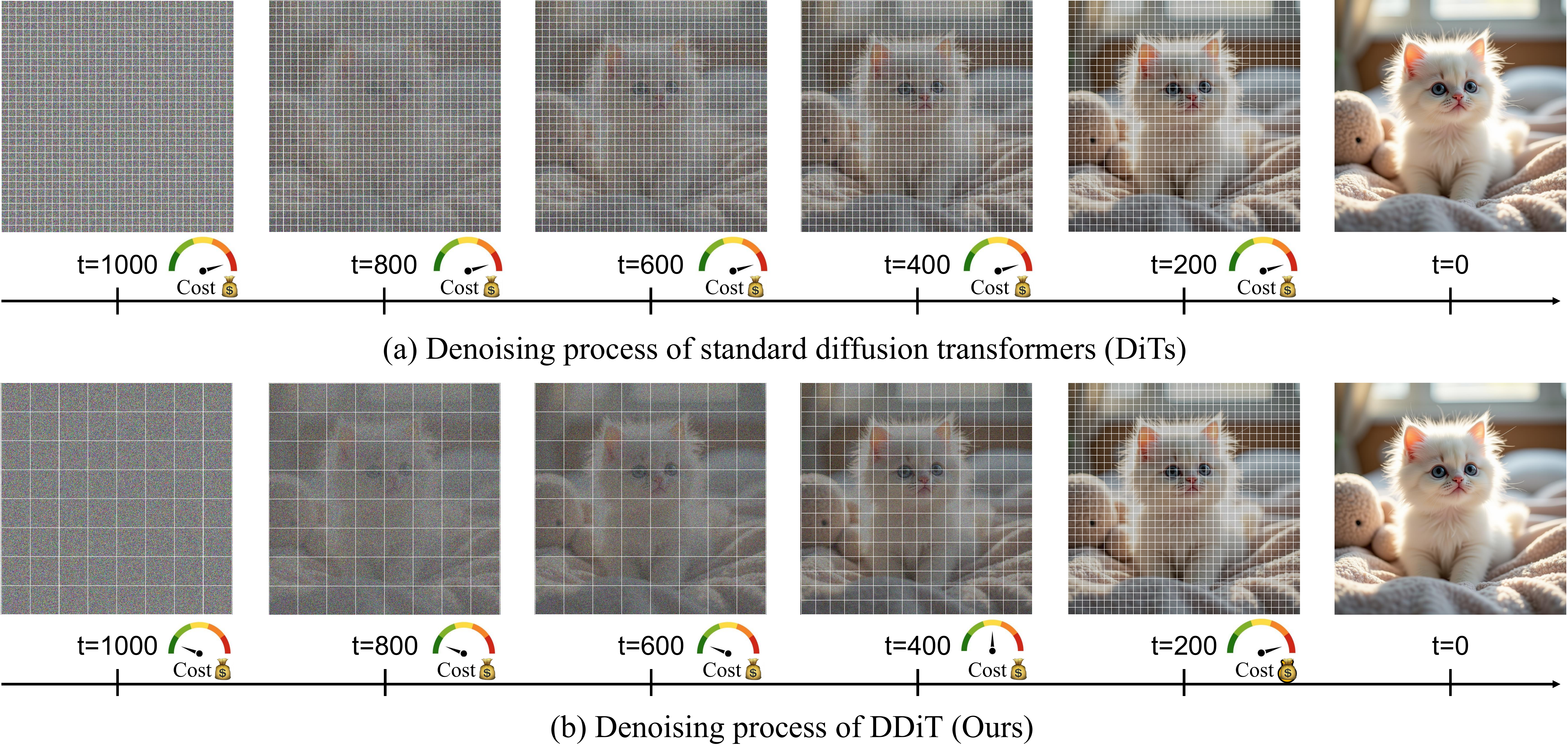}\vspace{-8pt}
   \caption{\footnotesize\textbf{Main idea: dynamic tokenization during denoising}. Current methods use the same patch size for \textit{all} denoising steps during inference time. 
   Instead, \textbf{DDiT} adapts the patch size at each timestep according to the latent complexity, allocating fewer tokens for certain timesteps and more tokens for certain others. While DiT divides VAE latents into patches, for illustrative purposes, we use a real image in pixel space.}
   \label{fig:fig2}\vspace{-1pt}
\end{figure*}

%% file: sec/3_method.tex
\section{Approach}
\label{approach}
Our goal is to achieve significant computational speedup at minimal loss of perceptual quality of image and video generations. We achieve this by \textit{dynamically} varying the patch size of a latent at each denoising timestep based on the complexity of the underlying latent manifold.
We first briefly introduce diffusion transformers (DiT)~\cite{peebles2023scalable} in Sec.~\ref{sec3.1}, motivate our approach to adapt DiT to dynamically process latent patches of different sizes in Sec.~\ref{sec3.2}, and finally
detail our novel approach to dynamically select the optimal latent patch size at every denoising step in (Sec.~\ref{sec3.3}). 

\input{sec/3_1}
\input{sec/3.2}

\input{sec/3.3}

%% file: sec/3_1.tex
\subsection{Preliminaries on Diffusion Transformers}
\label{sec3.1}

Owing to the flexibility and scalability of the transformer architecture~\cite{vaswani2017attention}, DiTs have achieved wide adoption in content generation. Built upon the Vision Transformer (ViT) architecture~\cite{dosovitskiy2020image}, DiTs operate in the latent space of a pre-trained variational autoencoder (VAE)~\cite{rombach2022high}. 
Briefly, given an input image $I$\footnote{For simplicity, we use image inputs, but our method is extensible to DiTs which process videos as we show in Sec.~\ref{sec:experiments}.}, it is first encoded by the VAE into a latent representation $\mathbf{z} \in \mathbb{R}^{H \times W \times C}$, where $H$, $W$, and $C$ denote the height, width, and channel dimensions of the latent feature map, respectively.
The input to the transformer-based diffusion model is this latent $\mathbf{z}$. 
During training, Gaussian noise is gradually added to $\mathbf{z}$, and DiT is optimized to predict and remove this noise~\cite{ho2020denoising}. 
During inference, the model starts from pure noise and iteratively denoises it over $T$ diffusion steps to recover a clean latent representation, which is then decoded by the VAE decoder to reconstruct the final image.
\input{figs_latex/fig2_ver2}

Note that the latent $z$ is pre-processed before feeding to the diffusion transformer. Specifically, $z$ is first divided into non-overlapping patches of size $p \times p$. Following this, each patch is \textit{tokenized} by passing through a patch embedding layer parameterized by weights $\mathbf{w}^{\text{emb}} \in \mathbb{R}^{p \times p \times C \times d}$ and bias $\mathbf{b}^{\text{emb}} \in \mathbb{R}^{d}$. This layer projects each patch into an embedding space of dimension $d$.
The resulting embeddings from each patch are then processed by $L$ stacked transformer blocks comprising a series of attention and feed-forward layers~\cite{vaswani2017attention}.
The attention mechanism learns to attend to relevant patches by computing pairwise dependencies among all $N = \frac{HW}{p^2}$ patches. Thus, attention operation has a computational complexity proportional to $\mathcal{O}(N^2)$.

Naturally, a smaller $p$ increases the number of tokens $N$, leading to an expensive attention operation, thereby higher computational cost per layer. Further, since denoising is an iterative operation, using a small patch size $p$ is even more computationally prohibitive. Moreover, prior studies have shown that not all denoising steps need the same level of granularity~\cite{mahajan2024prompting, patashnik2023localizing, stracke2025cleandift, tang2023emergent,kim2025revelio, wang2023not}. These factors motivate us to vary the patch size $p$ dynamically across timesteps to balance efficiency and generation quality.

%% file: figs_latex/fig2_ver2.tex
\begin{figure}[t]
  \centering
   \includegraphics[width=0.9\linewidth]{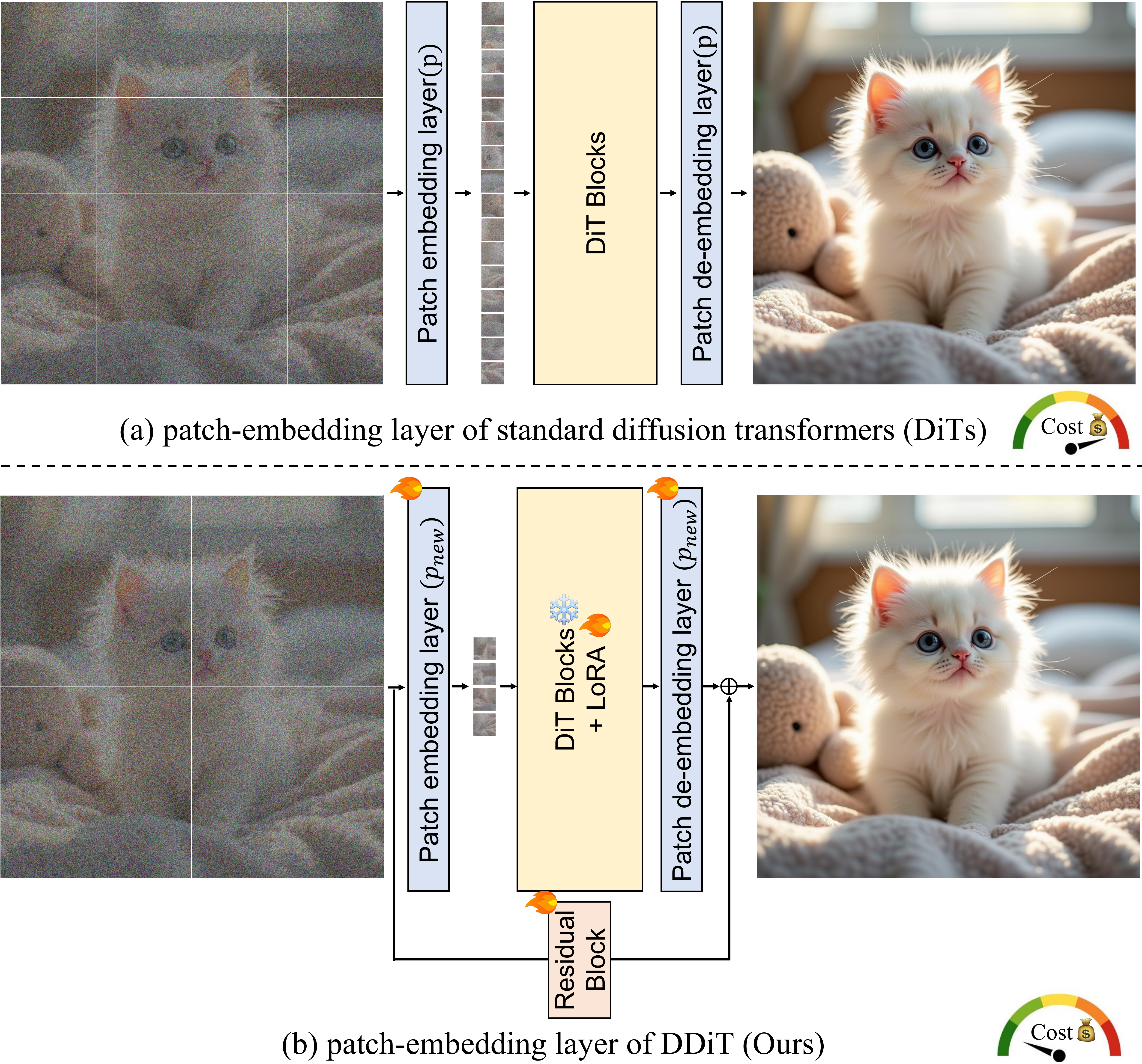}\vspace{-8pt}
   \caption{\footnotesize\textbf{Revised patch-embedding layer to support patches of varied resolutions.} We modify the standard patch-embedding layer, designed for a fixed patch size $p$, to additionally support patch sizes $p_{\text{new}}$. 
   }
   \vspace{-0.2in}
   \label{fig:architecture}
\end{figure}

%% file: sec/3.2.tex
\subsection{Dynamic Patching and Tokenization}
\label{sec3.2}
\input{figs_latex/speed_patch_size}
We aim to modify a pre-trained DiT to seamlessly operate under different patch sizes with minimal architectural modifications (Fig.~\ref{fig:architecture}). To this end, we adapt the patch embedding layer, originally operating on patch size $p$, to also handle new patch sizes $p_{\text{new}}$ and allow input latents of varying spatial resolutions. We define $p_{\text{new}}$ a positive integer multiple of $p$, \ie, $\{p, 2p, 4p, ...\}$. 

\noindent\textbf{Modifications to the patch embedding layer.} 
To generalize DiT to $p_{\text{new}}$, we introduce patch-specific embedding layers 
for each patch size we wish to support. 
Recall from Sec.~\ref{sec3.1} that $C$ denotes the number of latent channels and $d$ represents the embedding dimension. Let $\mathbf{w}^{\text{emb}}_{p_{\text{new}}} \in \mathbb{R}^{p_{\text{new}} \times p_{\text{new}} \times C \times d}$ and $\mathbf{b}^{\text{emb}}_{p_{\text{new}}} \in \mathbb{R}^{d}$ denote the weight matrix and bias vector of the patch embedding layer corresponding to ${p_{\text{new}}}$.
Each patch of size $p_{\text{new}}$ is linearly projected into an embedding of dimension $d$ using this newly added embedding layer. This results in a total of $N_{p_{\text{new}}} = \frac{HW}{p_{\text{new}}^2}$ patches. 
Since $N_{p_{\text{new}}}$ is smaller than $N$ by a factor of $(p_{\text{new}}/p)^2$, DiT now processes fewer patches and yields significant computational gains. As shown in Fig.~\ref{fig:speed_size}, increasing the patch size from $p$ to $2p$ yields a $~3 \times$ computational gain!

To minimize the training cost, we retain the base model originally trained on the latent patch size $p$ and introduce a Low-Rank Adaptation (LoRA) branch~\cite{hu2022lora} into \underline{each} transformer block in DiT.
This LoRA branch serves as an \textit{adaptive pathway} and enables the model to process patches of different sizes.
Additionally, as shown in Fig.~\ref{fig:architecture}, we add a residual connection from \textit{before} the patch embedding layer to \textit{after} the patch de-embedding block. This helps strike a balance between the base latent manifold and the new manifold being learnt by LoRA for $p_{\text{new}}$.

We reuse the learnt positional embeddings of the original patch size $p$ for $p_{\text{new}}$ by bilinearly interpolating them for the new patch size. We also introduce a learnable patch embedding  (a $d$-dimensional vector) added to all tokens akin to positional embeddings. This serves as a patch-size identifier and helps the model distinguish which patch size is being used at each timestep. At test time, we use the learned patch-size embedding as is. 

Finally, to distill the knowledge from the frozen base model to the LoRA-augmented model, we fine-tune the LoRA branch with a distillation loss.
Let $\epsilon_{\theta_L}$ and $\epsilon_{\theta_{T}}$ denote the predicted noise from the LoRA-fine-tuned and frozen base models respectively. The distillation loss is:
{\small
\begin{equation}
\mathcal{L} = || \epsilon_{\theta_L}(\mathbf{z}_t^{p_{\text{new}}}, t) - \epsilon_{\theta_{T}}(\mathbf{z}_t^p, t) ||_2^2.
\label{eq:distil_loss}
\end{equation}}
These minor architectural tweaks allow us to dynamically support larger patch sizes while maintaining the base model's perceptual output quality. We stress and empirically show in Sec.~\ref{sec:experiments} that these changes are seamlessly extensible to any diffusion-based image or video models. 
\input{figs_latex/std_illustrate}

%% file: figs_latex/speed_patch_size.tex
\begin{figure}[t]
  \centering
   \includegraphics[width=0.7\linewidth]{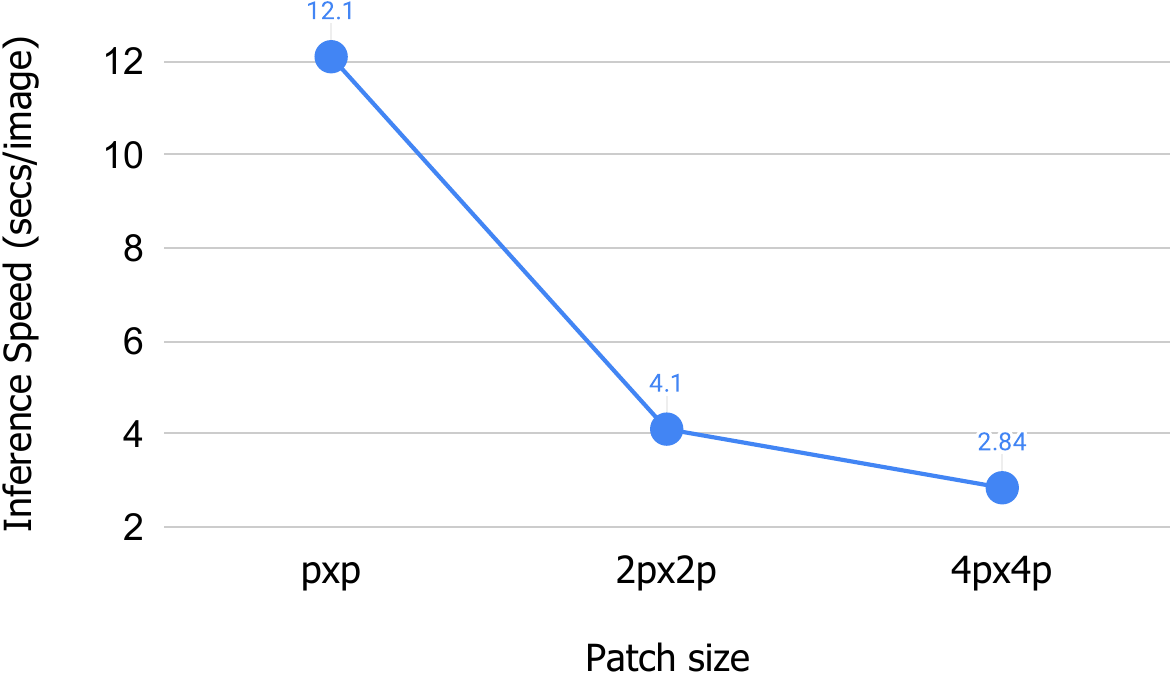}\vspace{-8pt}
   \caption{\footnotesize\textbf{Inference speed vs. patch size.} Inference speed measured over $50$ denoising steps for generating $1024\times1024$ images using FLUX-1.Dev~\cite{flux2023}, where every timestep uses a fixed patch size. As the patch size increases from $p$~$\rightarrow$$2p$~$\rightarrow$$4p$, the number of tokens decreases quadratically (4096~$\rightarrow$~1024~$\rightarrow$~256), resulting in approximately $3\times$ and $4\times$ faster inference for $2p$ and $4p$, respectively, compared to $p$.
 }\vspace{-8pt}
   \label{fig:speed_size}
\end{figure}

%% file: figs_latex/std_illustrate.tex
\begin{figure}[t]
  \centering
   \includegraphics[width=0.7\linewidth]{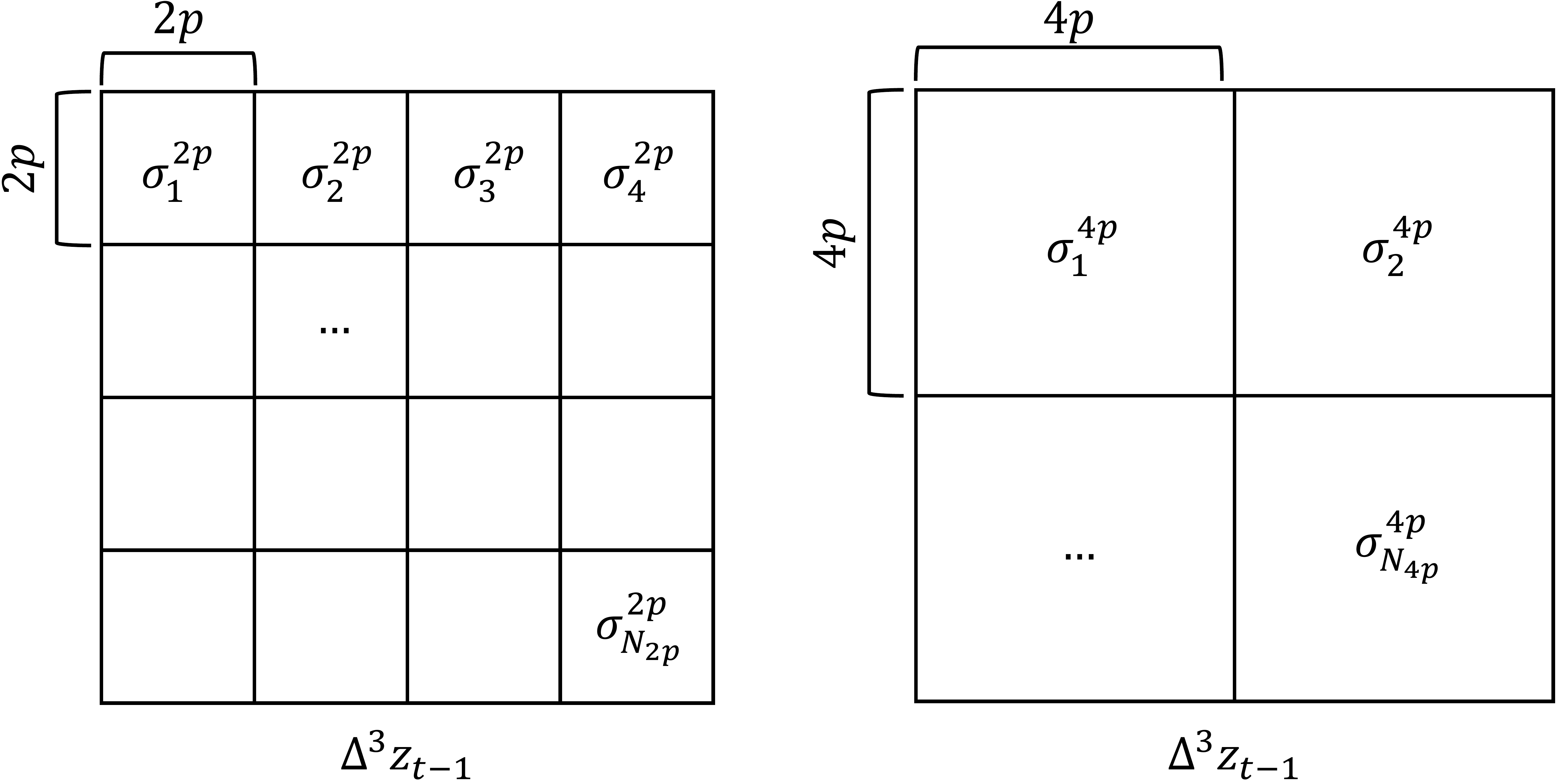}\vspace{-8pt}
    \caption{\footnotesize{Given $\Delta^{(3)}\mathbf{z}_{t-1}$, we divide it into patches of size $p_{i} \times p_{i}$, compute within-patch standard deviation $\boldsymbol\sigma_{t-1}^{p_i}$ of the acceleration.}}
    \vspace{-0.1in}
   \label{fig:std_illu}
\end{figure}

%% file: sec/3.3.tex
\subsection{Dynamic Patch Scheduling}
\label{sec3.3}

Now that we enabled processing of multiple size input patches, how do we learn \textit{when} to adapt to a larger patch-size (\ie, coarser token) and when to switch back to a smaller patch-size (\ie, fine-grained token)?
To this end, we introduce a dynamic patch scheduling mechanism to determine the appropriate patch size at each diffusion timestep. Since different timesteps correspond to different levels of generative detail~\cite{mahajan2024prompting, patashnik2023localizing, stracke2025cleandift, tang2023emergent, wang2023not, saichandran2025progressive}, selecting the proper patch size at each stage is crucial for maintaining both efficiency and quality. We hypothesize that: 
\begin{itemize}
    \item large patches may reasonably capture coarse scene structures without significant compromises of visual quality while yielding computational speedups.
    \item smaller patches may be pertinent to capture fine-grained details to retain all visual intricacies.
\end{itemize}
We automate this intuition in a highly light-weight manner and design a \textbf{training-free dynamic scheduler} that adaptively selects the patch size based on the \textit{rate of evolution of the latent representations} within a window of timesteps.  

\noindent \textbf{Latent evolution estimation.} We employ finite-difference approximations of increasing order to quantify how latent representations evolve during the denoising process.
Let $\mathbf{z}_t$ denote the latent at timestep $t$. The first-order finite difference captures the displacement of latent features between consecutive timesteps:
{\small
\begin{equation}
\Delta\mathbf{z}_{t} = \mathbf{z}_{t} - \mathbf{z}_{t+1}.
\end{equation}}
Similarly, the second-order difference describes the \textit{rate} of change of this displacement, representing the local velocity of the denoising trajectory, defined by
{\small
\begin{equation}
\Delta^{(2)}\mathbf{z}_{t-1} = \Delta\mathbf{z}_{t-1} - \Delta\mathbf{z}_{t}.
\end{equation}}
Finally, the third-order finite difference quantifies the variation in this velocity. This can be interpreted as a measure of \textit{acceleration} of a latent's evolution during denoising within a short temporal window.
{\small
\begin{equation}
\Delta^{(3)}\mathbf{z}_{t-1} = \Delta^{(2)}\mathbf{z}_{t-1} - \Delta^{(2)}\mathbf{z}_{t}
= 2 (\frac{\Delta\mathbf{z}_{t-1}+\Delta\mathbf{z}_{t+1}}{2} - \Delta\mathbf{z}_{t}),
\label{eq:latent_var}
\end{equation}}
\input{figs_latex/std_timestep2}

We hypothesize that if the acceleration is slow at a given timestep, there is a relatively minor difference in the underlying latent manifold in the local temporal window. On the other hand, \textbf{a high acceleration value suggests a larger difference in the structure of the underlying manifold.} We use this measure as a proxy to identify transition points where the generative process intrinsically shifts between generating coarse to fine structures or vice versa. 

Empirically, we find that the third-order difference captures this variation more effectively and remains more stable, while the first- and second-order differences fail to do so, likely because they capture relatively short-term temporal changes (Sec.~\ref{sec:4_analysis}). This observation is consistent with~\cite{ye2024training}, which shows that the difference between neighboring noise predictions is explicitly related to the third-order finite difference.  

\noindent \textbf{Spatial variance estimation.} We use latent $z_t$ in the above formulation for simplicity, but in practice, $z_t$ is always divided into
patches of size $p_{\text{new}} \times p_{\text{new}}$ (Sec.~\ref{sec3.2}). Our final task now is to select the right patch size at each latent manifold. This requires quantifying and aggregating the acceleration at which the latent patches evolve. Thus, we divide $z_{t-1}$ into patches of size $p_{i} \times p_{i}$, where $p_{i} \in p_{\text{new}}$. Then, we compute the \textbf{standard deviation} $\boldsymbol\sigma_{t-1}^{p_i}$ of the acceleration (defined in Eqn.~\ref{eq:latent_var}) within each patch (Fig.~\ref{fig:std_illu}). We hypothesize that if the per-(latent) pixel standard deviation within a latent patch is high, then the denoising process is focusing on generating finer-grained details. On the other hand, if the standard deviation is low, then the underlying evolving latent is smooth.
As shown in Fig.~\ref{fig:std_timestep2}, prompts with different levels of granularity exhibit distinct variance profiles across timesteps. 
\input{tables/t2i_sota}

Given $\boldsymbol\sigma_{t-1}^{p_i}$, our goal is to determine the appropriate patch size at each timestep.  
A straightforward way to do this is to aggregate $\boldsymbol\sigma_{t-1}^{p_i}$ by taking their mean across patches, which provides a simple measure of the overall latent variation at that timestep.
However, this fails to effectively capture the generative dynamics occurring at that timestep. For example, when generating an image containing both a uniform white background and a highly textured region, averaging might smoothen the higher standard deviation values, leading to the scheduler choosing larger patches and thus overlooking fine details in the textured area. To better capture such spatial heterogeneity in the underlying latent manifold, we instead take the $\rho$-th percentile of the per-patch variances, denoted as $\sigma_{t-1}^{^{p_i}, (\rho)}$. 
This percentile-based aggregation allows us to capture meaningful information across patches without averaging out important signals, while also avoiding bias toward a few high-variance outliers.

Concretely, we compare $\sigma_{t-1}^{^{p_i}, (\rho)}$ against a predefined variance threshold $\tau$. For each timestep, we select the largest patch size whose corresponding variance is below the threshold ($\tau$). If no such patch satisfies this condition, it defaults to the smallest patch size, which is $1$. We formulate this \textbf{patch size scheduling} as follows: 
{\small
\begin{equation}
    p_t =
    \begin{cases}
    \displaystyle \max(p_i), & \text{if } \sigma_{t-1}^{^{p_i}, (\rho)} < \tau, \\[6pt]
    \displaystyle $1$, & \text{otherwise}.
    \end{cases}
\label{eq:scheduling}
\end{equation}}
Controlling $\tau$ gives us explicit control over the speed: if users prefer faster generation, a higher $\tau$ can be selected; otherwise, a smaller $\tau$ can be used for higher quality with less speed gain. We select $\tau$ and $\rho$ empirically and balance generation stability and visual quality. 
In Sec.~\ref{sec:experiments}, we also show how such adaptive scheduling enables the generation process to allocate computational resources efficiently and strategically, while maintaining the overall generation fidelity. 

\input{figs_latex/zoom_in}

%% file: figs_latex/std_timestep2.tex
\begin{figure}[t]
  \centering
   \includegraphics[width=0.8\linewidth]{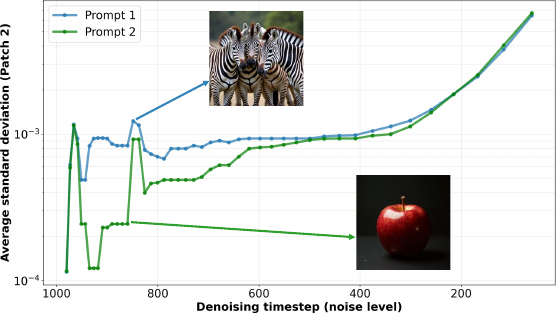}\vspace{-8pt}
    \caption{
    \footnotesize\textbf{Visualization of $\boldsymbol{\sigma}_{t-1}^{2p, (\rho)}$ for two prompts (log scale).} 
    \textit{Prompt 1: ``Several zebras are standing together behind a fence.''}
    \textit{Prompt 2: ``A simple red apple on a black background.''}
    Prompts requiring different levels of spatial granularity exhibit distinct $\boldsymbol{\sigma}_{t-1}^{2p, (\rho)}$ patterns across timesteps. 
    For the fine-grained zebra pattern, $\boldsymbol{\sigma}_{t-1}^{2p, (\rho)}$ remains higher, indicating higher detail sensitivity, 
    whereas for the simpler apple scene, $\boldsymbol{\sigma}_{t-1}^{2p, (\rho)}$ is lower, thus we can seamlessly use larger patch sizes during generation.
    }\vspace{-0.3in}
   \label{fig:std_timestep2}
\end{figure}

%% file: tables/t2i_sota.tex
\begin{table*}[ht]
\centering
\caption{\footnotesize\textbf{Quantitative comparison of text-to-image generation performance with state-of-the-art methods} on COCO, DrawBench, and PartiPrompts. 
If not specified, all results are reported using 50 inference steps by default. 
Each color (\colorbox{yellow!10}{\textbf{Yellow}}, \colorbox{cyan!10}{\textbf{Blue}}) indicates methods operating at similar inference speeds. 
As highlighted in \colorbox{cyan!10}{\textbf{Blue}}, our method achieves the best overall image quality, evidenced by the lowest FID scores, strong prompt alignment (CLIP and ImageReward), and high perceptual similarity (SSIM and LPIPS). \textbf{Bold}: best. \underline{Underline}: second-best. }\vspace{-8pt}
\small
\setlength{\tabcolsep}{4pt}
\resizebox{0.8\linewidth}{!}{
\begin{tabular}{lcccccccccccc}
\toprule
\multirow{2}{*}{\textbf{Model}} & \multirow{2}{*}{\textbf{Speed$\downarrow$} } &\multicolumn{2}{c}{\textbf{COCO}} & \multicolumn{4}{c}{\textbf{DrawBench}} & \multicolumn{2}{c}{\textbf{PartiPrompts}} \\
\cmidrule(lr){3-4} \cmidrule(lr){5-8}  \cmidrule(lr){9-10} 
&(secs/image) & \textbf{FID}$\downarrow$ & \textbf{CLIP}$\uparrow$ &
\textbf{CLIP}$\uparrow$ & \textbf{ImgR}$\uparrow$ & \textbf{SSIM}$\uparrow$  & \textbf{LPIPS}$\downarrow$ & \textbf{CLIP}$\uparrow$ & \textbf{ImgR}$\uparrow$  \\
\midrule
FLUX-1.Dev (50 steps)        & 12.0  & 33.07 & 0.314 & 0.3156 & 1.0291 &  --   &-- & 0.3197 &  1.192\\
\rowcolor{cyan!10} FLUX-1.Dev (28 steps)        & 6.72  & 33.35 & 0.312 & 0.3140 & 1.0107 & 0.739 $\pm$ 0.155   & 0.175 $\pm$ 0.126  & 0.3118 &  1.115 \\
\rowcolor{yellow!10} FLUX-1.Dev (15 steps)        & 3.6   & 34.02 & 0.311 & 0.3121 & 0.9865 & 0.591 $\pm$ 0.155  & 0.298 $\pm$ 0.128  & 0.3072 &  0.9613\\
\midrule
\rowcolor{cyan!10} TeaCache ($\delta=0.6$)        & 6.0   & 34.95 & 0.303 & 0.3071 & 0.9968 & \textbf{0.654 $\pm$ 0.145}  & \textbf{0.241 $\pm$ 0.114} & 0.3018 &  0.9701\\
 TeaCache ($\delta=0.8$)        & 5.33  & 35.57 & 0.303 & 0.3075 & 0.9780 & 0.612 $\pm$ 0.146  & 0.279 $\pm$ 0.114 & 0.2991 &  0.9699\\
\midrule
\rowcolor{cyan!10} TaylorSeer ($N=3, O=2)$      & 6.0   & 34.74 & 0.303 & 0.3085 & 0.9721 & 0.632 $\pm$ 0.173      & 0.271 $\pm$ 0.149  & 0.3142&  0.9813 \\
\rowcolor{yellow!10} TaylorSeer ($N=6, O=1)$           & 3.5   & 35.02 & 0.302 & 0.3040 & 0.9535 & 0.583 $\pm$ 0.130  & 0.284 $\pm$ 0.106 & 0.3004 &  0.9714\\
\midrule
\rowcolor{cyan!10} DDiT                  & 5.5   & \textbf{33.42} & \textbf{0.317} & \textbf{0.3136} & \textbf{1.0284} & \underline{0.635 $\pm$ 0.183}   & \underline{0.264 $\pm$ 0.190}& \textbf{0.3192} &  \textbf{1.189} \\
\rowcolor{yellow!10} DDiT + Teacache ($\delta=0.4$) & 3.4   & \underline{33.60} & \underline{0.315} & \underline{0.3117} & \underline{1.0182} & 0.592 $\pm$ 0.118   & 0.267 $\pm$ 0.120 & \underline{0.3172} &  \underline{1.062} \\
\bottomrule
\end{tabular}
}\vspace{-13pt}
\label{tab:t2i_sota}
\end{table*}

%% file: figs_latex/zoom_in.tex
\begin{figure}[t]
\begin{center}
\centerline{\includegraphics[width=0.9\linewidth]{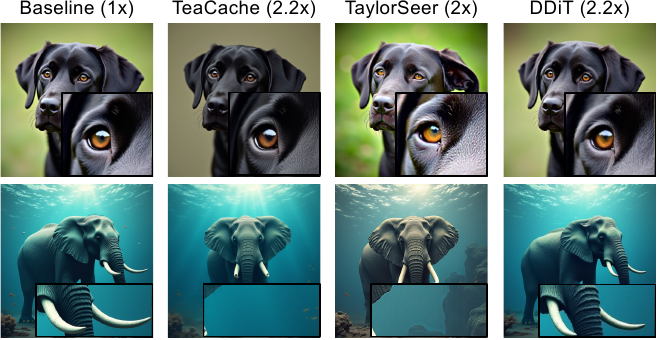}}\vspace{-8pt}
\caption{\footnotesize{
Qualitative comparisons with the base model~\cite{flux2023}, TeaCache~\cite{liu2025timestep}, TaylorSeer~\cite{liu2025reusing}, and DDiT under similar speedups on DrawBench. 
\textbf{DDiT} effectively preserves fine-grained details, pose, spatial layout, and overall color distribution of the generated images.
}}\vspace{-0.4in}
\label{fig:zoom_in}
\end{center}
\end{figure}

%% file: sec/4_experiments.tex
\section{Experiments}\label{sec:experiments}

\input{sec/4_setup}
\input{figs_latex/qual}

\input{figs_latex/fig_video}

\input{sec/4_t2i}

\input{tables/vbench_score}

\input{sec/4_t2v}

\input{tables/effect_nth_order}
\input{figs_latex/effect_schedule}

\input{sec/4_analysis}
\input{tables/effect_threshold}

%% file: sec/4_setup.tex
\subsection{Setup}
\noindent\textbf{Implementation details.}  
We use FLUX-1.dev~\cite{flux2023} and Wan-2.1 1.3B~\cite{wan2025} as base models for the text-to-image (T2I) and text-to-video (T2V) experiments, respectively.  
To support new patch sizes $p_{\text{new}}$, we introduce corresponding patch embedding and de-embedding layers for the patchify operation, along with corresponding patch positional embeddings.
We also add LoRA parameters~\cite{hu2022lora} with a rank of $32$ into the feed-forward layers of each transformer block and a single residual block, which are then fine-tuned along with the newly introduced components.
For both T2I and T2V models, we support patch sizes $p_{\text{new}} = 2p,4p$, although our method in principle can be extended to any size patches.
For both T2I and T2V tasks, we use 50 inference steps for comparison, but our method can be applied with any number of inference steps.
The T2I model is finetuned on the T2I-2M dataset~\cite{jackyhate2024}, a synthetic dataset generated using the base model, and the T2V model is trained on synthetic videos generated by the base model using prompts from the Vchitect-T2V-Dataverse~\cite{fan2025vchitect}.  
We use Prodigy~\cite{mishchenko2023prodigy}, an optimizer that automatically finds the optimal learning rate without requiring manual tuning, with a learning rate of 1.0 for T2I. 
For the T2V model, we employ AdamW~\cite{loshchilov2017decoupled} with a learning rate of $1\times10^{-4}$.
We initialize the patch-embedding weights using the pseudo-inverse of the bilinear-interpolation projection following~\cite{beyer2023flexivit}, which helps preserve the base model’s functional behavior.
To balance visual fidelity and computational efficiency, we set $\tau = 0.001$ and $\rho=0.4$ for all experiments. 

\noindent\textbf{Evaluation setup.}  
For evaluation, we generate images at a resolution of $1024 \times 1024$ using 50 inference steps and a guidance scale of 3.5 for the text-to-image task, and videos at $480 \times 832$ resolution with 81 frames using 50 inference steps for the text-to-video task.  
As commonly done, for text-to-image evaluation, we use the COCO dataset~\cite{lin2014microsoft} to compute CLIP~\cite{hessel2021clipscore, radford2021learning} and FID~\cite{heusel2017gans} scores against real images, assessing overall visual quality.  
We additionally evaluate on DrawBench~\cite{saharia2022photorealistic} and PartiPrompts~\cite{yu2022scaling} datasets using CLIP score and ImageReward~\cite{xu2023imagereward} to measure text–image alignment, and SSIM~\cite{wang2004image} and LPIPS~\cite{zhang2018unreasonable} to assess structural similarity with the base model.  
For text-to-video evaluation, we adopt VBench~\cite{huang2024vbench} and follow the evaluation protocol proposed in their work.

%% file: figs_latex/qual.tex
\begin{figure}[t]
  \centering
  \includegraphics[width=0.7\linewidth]{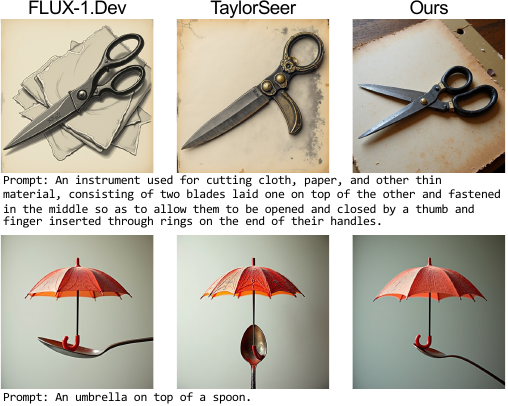}\vspace{-0.1in}
   \caption{\footnotesize\textbf{Qualitative comparison on DrawBench} with the baseline and TaylorSeer~\cite{liu2025reusing}.  Our method remains robust even for complex prompts that require a deeper understanding of semantic content. }\vspace{-9pt}
   \label{fig:qual}
\end{figure}

%% file: figs_latex/fig_video.tex
\begin{figure}[t]
  \centering
   \includegraphics[width=\linewidth]{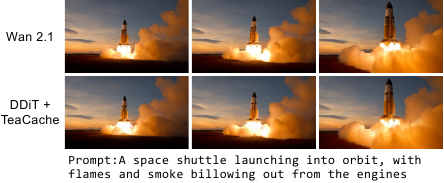}\vspace{-8pt}
    \caption{
\footnotesize\textbf{Qualitative comparison of text-to-video generation between DDiT and the baseline.} 
DDiT produces videos with comparable visual quality to the baseline while achieving significant speedup.
}\label{fig:video}\vspace{-8pt}
\end{figure}

%% file: sec/4_t2i.tex
\subsection{Text-to-Image Generation}

We first evaluate the effectiveness of our approach on the text-to-image (T2I) generation task and compare it with state-of-the-art acceleration methods. Our evaluation focuses on measuring both efficiency and perceptual quality, as reducing inference cost often leads to a loss in fine-grained visual details or degraded text–image alignment.  
We use FLUX-1.dev~\cite{flux2023} as the base model and vary the number of inference steps to simulate different computational budgets. Our approach is compared against TeaCache~\cite{liu2025timestep} and TaylorSeer~\cite{liu2025reusing}, two state-of-the-art caching-based acceleration methods, under multiple configurations to ensure a fair comparison. 

As shown in Table~\ref{tab:t2i_sota} and Fig.~\ref{fig:zoom_in}, our method achieves substantial improvements in inference speed while maintaining high generation quality.  Compared to the base model, our approach achieves comparable FID (only a 0.35 difference) and CLIP scores, while delivering a $\mathbf{2.18\times}$ speedup. This demonstrates that dynamically adjusting patch sizes across denoising steps enables more efficient computation without compromising perceptual fidelity. Under similar inference speeds (rows 4, 6, and 8), our method consistently outperforms prior approaches. 
As shown in Fig.~\ref{fig:qual}, our model seamlessly handles complex cases that require a deeper understanding of the semantic content of the prompt.
Moreover, our method preserves the overall perceptual similarity to the base model’s output while substantially reducing inference cost.

\noindent \textbf{Combining with TeaCache}~\cite{liu2025timestep}: 
Furthermore, our approach is complementary to existing acceleration strategies such as caching.  
When combined with TeaCache (row 9), our method achieves a $\mathbf{3.52\times}$ speedup over the baseline! This \textbf{surpasses all existing state-of-the-art approaches} in both efficiency and generation quality.  
These results confirm that our dynamic patch scheduling strategy effectively balances computation and quality, offering a simple yet powerful mechanism for efficient diffusion generation.

%% file: tables/vbench_score.tex
\begin{table}[t]
\centering
\caption{\footnotesize\textbf{Quantitative results on V-Bench~\cite{huang2024vbench}.} 
Comparison of DDiT under different threshold settings ($\tau$) and its combination with TeaCache~\cite{liu2025timestep}. 
}\vspace{-8pt}
\resizebox{0.9\linewidth}{!}{
\begin{tabular}{lcc}
\toprule
\textbf{Model} & \textbf{Speed ($\uparrow$)} & \textbf{VBench ($\uparrow$)} \\
\midrule
Wan-2.1 (Baseline) & 1.0$\times$ & 81.24 \\ \midrule
Ours ($\tau{=}0.004$) & 1.6$\times$ & 81.17 \\
Ours ($\tau{=}0.001$) & 2.1$\times$ & 80.97 \\
Ours ($\tau{=}0.001$) + TeaCache ($\delta{=}0.05$) & 3.2$\times$ & 80.53 \\
\bottomrule
\end{tabular}
}
\label{tab:wan_vbench}\vspace{-8pt}
\end{table}

%% file: sec/4_t2v.tex
\subsection{Text-to-Video Generation}

We further evaluate the effectiveness of our method in the text-to-video (T2V) generation setting and compare it with the base model~\cite{wan2025}. 
Our method dynamically adjusts the patch size across denoising steps, allowing the model to allocate computation adaptively according to the complexity of spatial structures.  

As shown in Table~\ref{tab:wan_vbench}, our approach significantly reduces inference time while maintaining competitive video quality, as reflected by the VBench score~\cite{huang2024vbench}.  
Qualitative results in Fig.~\ref{fig:video} show that our method preserves motion consistency and fine-grained frame details even at accelerated inference speeds. Additional results are in Appendix.

%% file: tables/effect_nth_order.tex
\begin{table}[t]
\centering
\caption{\footnotesize\textbf{Effect of the $n$-th order difference on generation quality.}
Higher-order terms capture more informative temporal dynamics, improving both FID and CLIP scores.  
The third-order term ($n=3$) achieves the best overall performance.}\vspace{-8pt}
\small
\resizebox{0.6\linewidth}{!}{
\begin{tabular}{lccc}
\toprule
\textbf{Method} & \textbf{FID↓} & \textbf{CLIP↑} & \textbf{ImageReward↑} \\
\midrule
DDiT ($n=1$) & 34.71 & 0.2927 & 0.9782 \\
DDiT ($n=2$) & 34.28 & 0.3082 & 1.0128 \\
DDiT ($n=3$) & \textbf{33.42} & \textbf{0.3136} & \textbf{1.0284} \\
\bottomrule
\end{tabular}}\vspace{-8pt}
\label{tab:order_effect}
\end{table}

%% file: figs_latex/effect_schedule.tex
\begin{figure}[t]
  \centering
  \includegraphics[width=0.8\linewidth]{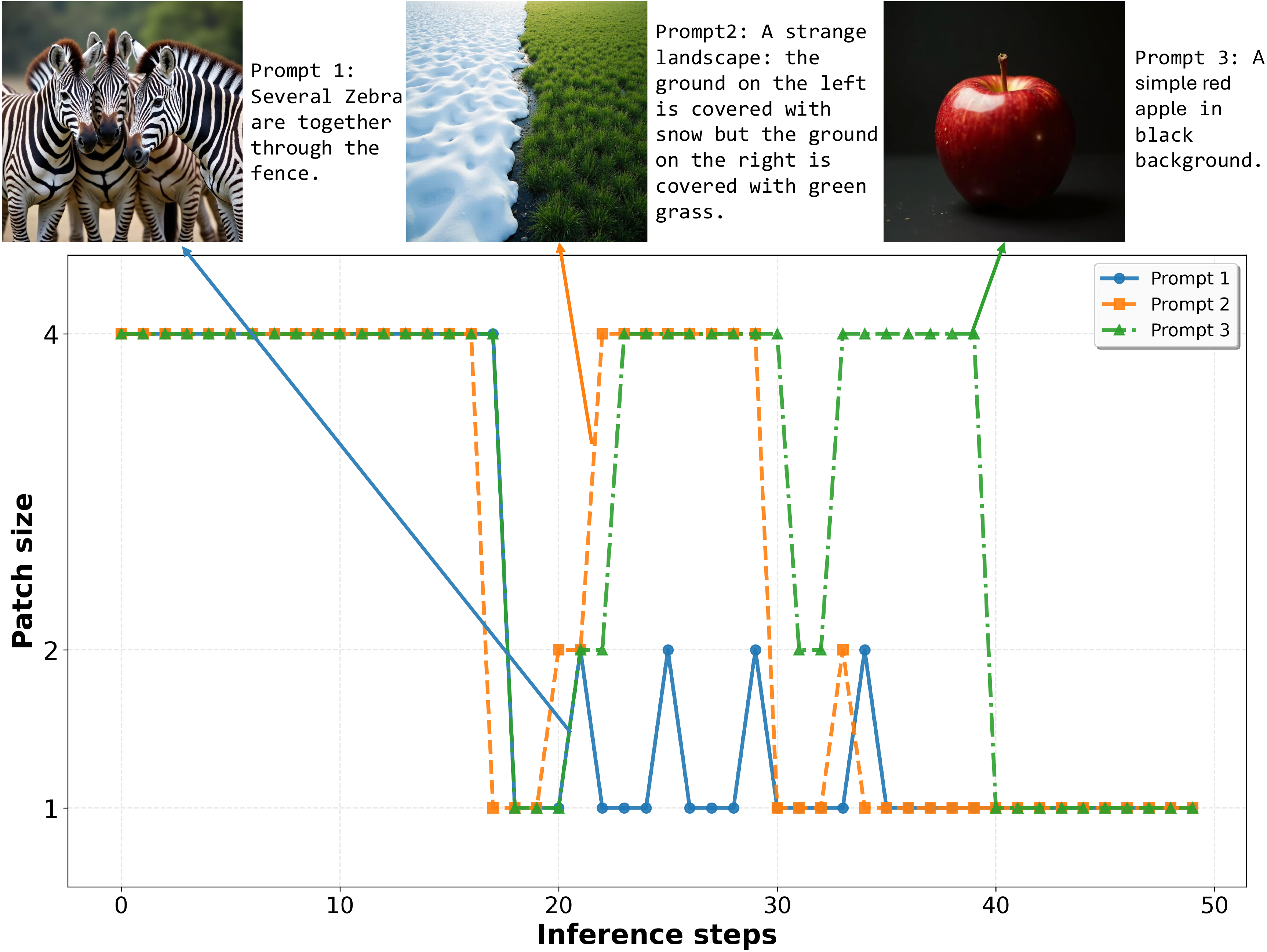}\vspace{-8pt}
    \caption{\footnotesize\textbf{Sample patch schedules} for 3 different prompts. Our dynamic patch scheduler seamlessly adapts to each prompt’s complexity and detail, thereby allocating more computation (aka higher percentage of smaller patch sizes) to images with highly detailed textures compared to simpler ones, thereby balancing efficiency and visual quality.}
    \vspace{-0.2in}
  \label{fig:effect_schedule}
\end{figure}

%% file: sec/4_analysis.tex
\subsection{Analysis}\label{sec:4_analysis}
In this section, we conduct an extensive analysis of our method to better understand our method.

\noindent\textbf{Effect of speedup on visual quality: a user study.} 
To assess whether humans can distinguish between DDiT and baseline generations, we conduct a user study on visual preference.
Raters were shown image pairs (DDiT vs. baseline) presented side by side in random order and asked to select the image with higher visual quality.
We find that generations from DDiT are visually as pleasing and photo-realistic as DiT $\mathbf{61\%}$ of the time, while DiT generations are preferred over DDiT 22\% of the time. Surprisingly, we find that DDiT generations are preferred over DiT baseline 17\% times, even though this was not our main goal. 
These results clearly demonstrate that DDiT achieves visual quality on par with the baseline while providing substantial speedup.

\noindent\textbf{Effect of $n$-th order difference equation.}  
Table~\ref{tab:order_effect} shows the impact of employing different $n$-th order terms in our latent variation estimation.  
As $n$ increases, both FID and CLIP scores consistently improve, suggesting that higher-order differences capture richer and more informative temporal dynamics of the latent space throughout the denoising process.  
In particular, the third-order term ($n=3$) achieves the best overall performance, producing the lowest FID and the highest CLIP and ImageReward scores.

\noindent\textbf{Effect of the patch schedule across different prompts.} We examine how our dynamic patch scheduling mechanism adapts to different text prompts with varying levels of complexity.  
As illustrated in Fig.~\ref{fig:effect_schedule}, our method automatically adjusts the patch schedule based on the semantic and structural richness of each prompt, effectively reallocating computational resources throughout the denoising process.  
For prompts describing complex scenes that involve fine-grained textures, the scheduler assigns more denoising steps with finer patches to capture detailed visual information. Conversely, for simpler prompts that depict minimal structures or uniform backgrounds, the model adaptively switches to coarser patches, thereby reducing redundant computation and accelerating inference.  
This adaptive behavior allows the model to balance efficiency and quality on a per-prompt basis, ensuring that computational effort is concentrated where it contributes most to perceptual fidelity. Overall, this demonstrates that our patch scheduling strategy not only accelerates generation but also enables content-aware allocation of computation, leading to improved scalability and robustness across diverse prompt distributions.

\noindent\textbf{Effect of the threshold on patch scheduling.}  
We analyze the impact of varying the threshold $\tau$ used in our patch scheduling mechanism, which determines when to switch between coarse and fine patch sizes during denoising. As shown in Table~\ref{tab:threshold_drawbench}, increasing $\tau$ results in slightly lower visual quality across all metrics, including FID, CLIP, and ImageReward scores. 
This trend can be attributed to the patch size scheduler becoming less sensitive to temporally local variations of the latent manifold at higher thresholds, leading to the premature selection of coarser patches and loss of fine-grained details. Nevertheless, the degradation remains small, confirming the robustness of our scheduling strategy.  
To balance visual fidelity and computational efficiency, we set $\tau = 0.001$ for all experiments.

%% file: tables/effect_threshold.tex
\begin{table}[t]
\centering
\caption{\footnotesize\textbf{Effect of the threshold $\tau$ on DrawBench.} 
Higher $\tau$ values yield faster inference at very mild dip in generation quality.}\vspace{-8pt}
\label{tab:threshold_drawbench}
\resizebox{0.7\columnwidth}{!}{
\begin{tabular}{lccc}
\toprule
\textbf{Method} & \textbf{Speed ($\times$)} & \textbf{CLIP} & \textbf{ImageReward} \\
\midrule
DDiT ($\tau{=}0.004$) & 1.88 & 0.3148 & 1.0271  \\
DDiT ($\tau{=}0.001$) & 2.18 & 0.3136 & 1.0284 \\
DDiT ($\tau{=}0.01$)  & 3.52 & 0.3082 & 1.0124 \\
\bottomrule
\end{tabular}
}\vspace{-12pt}
\end{table}

%% file: sec/5_conclusion.tex
\section{Conclusion and Future Work}
We present an intuitive and highly-computationally efficient method, {\modelname}, to adapt diffusion transformers to patches of different sizes during denoising while maintaining visual quality. {\modelname} demonstrates a critical insight: not all timesteps require the underlying latent space to be equally fine-grained. Building on this insight, we dynamically select the optimal patch size at every timestep and achieve significant computational gains, with no loss in perceptual visual quality. Our approach requires just adding a simple plug-and-play LoRA adapter to make the patch-embedding (and de-embedding) blocks amenable to varied input patch sizes. This minimal architectural tweak allows any DiT-based model to benefit from fast inference. Notably, it can also be applied to long-video generation, allowing the model to generate longer videos with the same amount of compute.
In our current design, for a given timestep, we use a fixed patch-size, but vary patch-sizes across timesteps. A natural future research would involve investigating varied patch sizes \textit{within} a given timestep, for further efficiency.